%% file: main.tex
\documentclass{article} 
\usepackage{iclr2022_conference,times}

\input{math_commands.tex}

\usepackage{tabularx}
\usepackage{graphicx}
\usepackage{hyperref}
\usepackage{url}
\usepackage[section]{placeins} 

\title{Learning latent representations for operational nitrogen response rate prediction}

\author{Christos Pylianidis, Ioannis N. Athanasiadis\\
Wageningen University\\
Wageningen, 6708 PB, The Netherlands \\
\texttt{\{christos.pylianidis,ioannis.athanasiadis\}@wur.nl} \\
}

\iclrfinalcopy 
\begin{document}

\maketitle

\begin{abstract}
Learning latent representations has aided operational decision-making in several disciplines. Its advantages include uncovering hidden interactions in data and automating procedures which were performed manually in the past. Representation learning is also being adopted by earth and environmental sciences. However, there are still subfields that depend on manual feature engineering based on expert knowledge and the use of algorithms which do not utilize the latent space. Relying on those techniques can inhibit operational decision-making since they impose data constraints and inhibit automation. In this work, we adopt a case study for nitrogen response rate prediction and examine if representation learning can be used for operational use. We compare a Multilayer Perceptron, an Autoencoder, and a dual-head Autoencoder with a reference Random Forest model for nitrogen response rate prediction. To bring the predictions closer to an operational setting we assume absence of future weather data, and we are evaluating the models using error metrics and a domain-derived error threshold. The results show that learning latent representations can provide operational nitrogen response rate predictions by offering performance equal and sometimes better than the reference model.
\end{abstract}

\section{Introduction}

Latent representation learning has been adopted in several disciplines to extract and handle hidden interactions between the input variables allowing for more informed decisions. In geosciences, representation learning algorithms emerge \citep{Jean2018Tile2Vec:Data} that perform visual analogies in the latent space, similar to how Word2vec can be leveraged to learn how words appear in similar contexts. In medicine, latent representation learning is used \citep{Zhou2019LatentData} to work with incomplete multi-modality data to learn independent representations for the prediction of Alzheimer's appearance. In biology, latent representations are used to model unmeasured quantities like pain and stress \citep{Kopf2021LatentMedicine}. Representation learning has also found its way to the earth and environmental sciences. Examples include learning better representations of 2D coordinates \citep{Mai2022Sphere2Vec:Predictions}, and extracting unknown basin characteristics \citep{Ghosh2021Knowledge-guidedCharacteristics}. However, it has been observed \citep{Neumann2019In-domainSensing} that this is not the case for several subfields, where practitioners prefer to use features based on expert knowledge and already proven algorithms that do not explore the latent space. This creates missed opportunities to examine whether improved predictive performance can be achieved, or new interactions to be found, or even to automate prediction pipelines. A representative case of such a missed opportunity is with estimating nitrogen application for fertilization purposes. 

Nitrogen is the nutrient that crops and pasture draw from the soil in the greatest quantities \citep{Rivai2021NitrogenSeedlings} and thus it becomes a growth-limiting factor \citep{Zhou2019ImagingTreatments}. Nitrogen deficiency has been associated with low yields \citep{Zhang2015TOND1Rice}, and pastures in several countries suffer from it \citep{Rotz2005Whole-farmAgriculture, Whitehead1995GrasslandNitrogen}. Farmers apply nitrogen-containing fertilizer to increase pasture growth rates but environmental concerns rise as nitrogen has been linked to soil\citep{Han2015ThePlants}, freshwater and atmosphere pollution\citep{Zhang2015ManagingDevelopment}. Subsequently, agricultural practitioners are asked to control nitrogen application with precise doses based on nitrogen response rate\footnote{Amount of extra $kg$ of yield for every $kg$ of nitrogen applied ($kg_{yield}/ha/kg_{nitrogen}$)}(NRR). To control nitrogen application, research is being directed towards modern systems like digital twins \citep{Nasirahmadi2022TowardParadigm} which can aid decision support through automation and data integration. However, digital twins require components, such as process-based and machine learning (ML) models, that are able to predict NRR across several months in the future to be considered operational. Process-based models that can calculate NRR exist but they are of limited use as the weather months after nitrogen application is unknown yet required to run the model. Also, while NRR observations exist from experiments, they are sparse and not enough to train ML models.

In a recent study \citep{Pylianidis2022Simulation-assistedTwins}, we presented a methodology to tackle these data-related problems by training ML models based on process-based model output. We predicted pasture NRR two months ahead of the prediction date, assuming an absence of intermediate weather data. However, we performed common practices of environmental sciences like selecting features solely based on expert knowledge, averaging weather variables, and feeding all those to Random Forest (RF) \citep{Breiman2001RandomForests}. Hence, the latent space of data was not explored, and it was left unchecked if we could achieve similar performance with higher resolution data by learning the latent space. That would be important to examine since methods that learn the latent space have shown to perform equally or better than approaches that do not, as they may capture interactions that are not yet understood. Also, it would promote automation in systems like digital twins by removing the step of manual feature extraction. In this work, we are going to treat this study as a stepping stone, as it proved that we can have accurate NRR predictions in limited data settings, in a situation where an ML model and a process-based model alone were not operational.  

Here, we perform a systematic comparison of different architectures to learn the latent space of a synthetic dataset for NRR prediction. We adopt the case study and data provided by \citep{Pylianidis2022Simulation-assistedTwins} and we use RF as a reference for comparing the performance of the architectures. We learn the latent representations of the inputs/outputs of a process-based model and predict NRR with a Multilayer Perceptron (MLP), an autoencoder (AE), and a dual-head autoencoder (DAE). We perform multiple runs for each architecture as well as RF to verify the robustness of each model. We then evaluate the results using error metrics as well as a domain-derived error threshold.

\section{Materials and Methods}

\subsection{Case study \& Data generation}
The case study was concerned with finding the pasture NRR for two sites (Fig.~\ref{fig:NZ_sites}) in New Zealand. The prediction target was the NRR of pasture dry matter grown in the two months after fertilizer application. Data generation was performed with APSIM \citep{Holzworth2014APSIMSimulation}. The simulation parameters of APSIM covered conditions that are known to affect pasture growth. The full factorial \citep{Antony20146Designs} of those parameters was created and put to APSIM. The range of each parameter can be seen in Table~\ref{tab:apsim_inputs}.

\subsection{Data preprocessing}
The generated data were processed to form a regression problem. The target variable was the NRR and the input variables were the weather, fertilizer amount, fertilization month, irrigation and a subset of biophysical variables produced by APSIM. From the generated daily data, only the data within the first 28 days prior to fertilization were preserved because pasture is supposed to 'lose memory' of past conditions after that time frame. \textbf{Weather data after that these 28 days were also discarded as they would be unavailable in operational conditions}. The remaining data were split into $67.5\%$ training, $12.5\%$ validation, and $20\%$ test sets, based on years, to avoid information leakage during later processing stages. The validation set included the years [1979, 1987, 1999, 2007], the training set years [1979-2010] excluding the validation years, and the test set [2011-2018].

\subsection{Architectures}

\subsubsection{Multilayer Perceptron}
An MLP was put in the comparison to examine how its latent space learning capabilities compared with learning compressed representations of an AE. The loss was given by equation~\ref{eq:1}. The network topology can be seen in Fig.~\ref{fig:mlp_topology}. Training parameters can be found in Appendix~\ref{tuning_training}.

\begin{figure}[h]
\begin{center}
\includegraphics[width=0.45\linewidth]{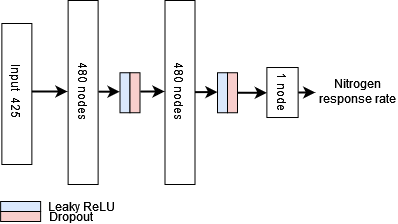}
\end{center}
\caption{The topology of the MLP.}
\label{fig:mlp_topology}
\end{figure}

\subsubsection{Autoencoder}
An AE was selected to create a compressed representation of the input variables. The AE included skip connections similarly to \citep{Li2018AutoencoderEstimation} from the encoder to the decoder to lessen degradation \citep{He2016DeepRecognition}. The reconstruction loss was given by equation~\ref{eq:2}. After training the AE, the decoder was removed and replaced by an MLP. Training was performed again for the MLP (with loss given by equation~\ref{eq:1}, and a frozen encoder) to learn to predict NRR. The autoencoder topology can be seen inside the dashed line of Fig.~\ref{fig:autoencoder_topology}. 

\noindent\begin{tabularx}{\textwidth}{@{}XX@{}}
  \begin{equation}
  L_{nrr}=\frac{1}{N}\sum\limits_{i=1}^{N}(y_i-\hat{y_i})^2
    \label{eq:1}
  \end{equation} &
  \begin{equation}
  L_{rec}=\frac{1}{N}\sum\limits_{i=1}^{N}\sum\limits_{j=1}^{425}(y_{ij}-\hat{y_{ij}})^2
    \label{eq:2}
  \end{equation}
\end{tabularx}

where $N=batch \ size$.

\subsubsection{Dual-head autoencoder}
\begin{figure}[h]
\begin{center}
\includegraphics[width=0.8\linewidth]{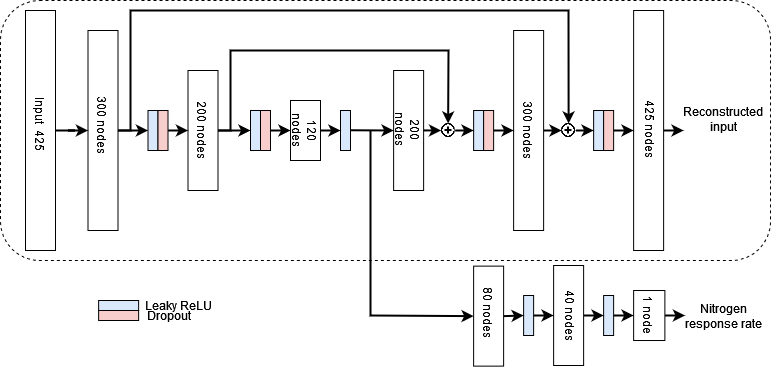}
\end{center}
\caption{The topologies of the AE (inside the dashed border), and the DAE (altogether).}
\label{fig:autoencoder_topology}
\end{figure}

The encoder and decoder parts were the same as of the 'simple' AE. The addition was that the compressed representation was then directed to an MLP which carried out the NRR prediction task. The network topology can be seen in Fig.~\ref{fig:autoencoder_topology}. Both the AE and the MLP were trained simultaneously, with the total loss being the summation of the equations~\ref{eq:1},~\ref{eq:2}.

\subsection{Evaluation}
The performance of the different architectures was compared using the \textit{mean absolute error} (MAE), the variance learned from the latent representations using $R^2$, and the standard deviation of the predictions. Also, the predictive capacity of the models was assessed using a domain-derived error threshold of 5 $kg_{yield}/ha/kg_{nitrogen}$. Prediction residuals systematically above that threshold constituted a model incapable for operational use. Each architecture, as well as RF, were ran 5 times with different seeds to verify the robustness of the results.

\section{Results}

In Table~\ref{tab:metrics}, we see the error metrics for each architecture and location aggregated over the runs. RF has the lowest error and highest explained variance for both locations. AE has the largest error and lowest $R^2$. DAE has the lowest errors among the architectures with just a slight edge over MLP. Regarding the standard deviations of the predictions, AE has the lowest deviation and DAE the highest.

\begin{table}[htp]
\caption{Error metrics for each architecture and RF aggregated over the runs. $\sigma$ refers to the standard deviation of the predictions of the runs.}
\label{tab:metrics}
\begin{center}
\begin{tabular}{l|ccc|ccc|ccc|ccc}
\multicolumn{1}{c}{\bf }  &\multicolumn{3}{c}{\bf RF} &\multicolumn{3}{c}{\bf MLP} &\multicolumn{3}{c}{\bf AE} &\multicolumn{3}{c}{\bf DAE}\\
\multicolumn{1}{c}{\bf }  &\multicolumn{1}{c}{ MAE} &\multicolumn{1}{c}{\bf $R^2$} &\multicolumn{1}{c}{\bf $\sigma$} &\multicolumn{1}{c}{ MAE} &\multicolumn{1}{c}{\bf $R^2$} &\multicolumn{1}{c}{\bf $\sigma$} &\multicolumn{1}{c}{ MAE} &\multicolumn{1}{c}{\bf $R^2$} &\multicolumn{1}{c}{\bf $\sigma$} &\multicolumn{1}{c}{ MAE} &\multicolumn{1}{c}{\bf $R^2$} &\multicolumn{1}{c}{\bf $\sigma$}
\\\hline\\
    Waiotu   & 1.55 & 0.68 & 3.53 & 1.85 & 0.62 & 3.63 & 2.26 & 0.45 & 3.34 & 1.72 & 0.65 & 3.62 \\
    Mahana   & 1.87 & 0.61 & 4.16 & 2.19 & 0.53 & 4.57 & 2.72 & 0.38 & 4.02 & 2.07 & 0.5 & 4.9 \\
\end{tabular}
\end{center}
\end{table}

\begin{figure}[h]
\begin{center}
\includegraphics[width=\linewidth]{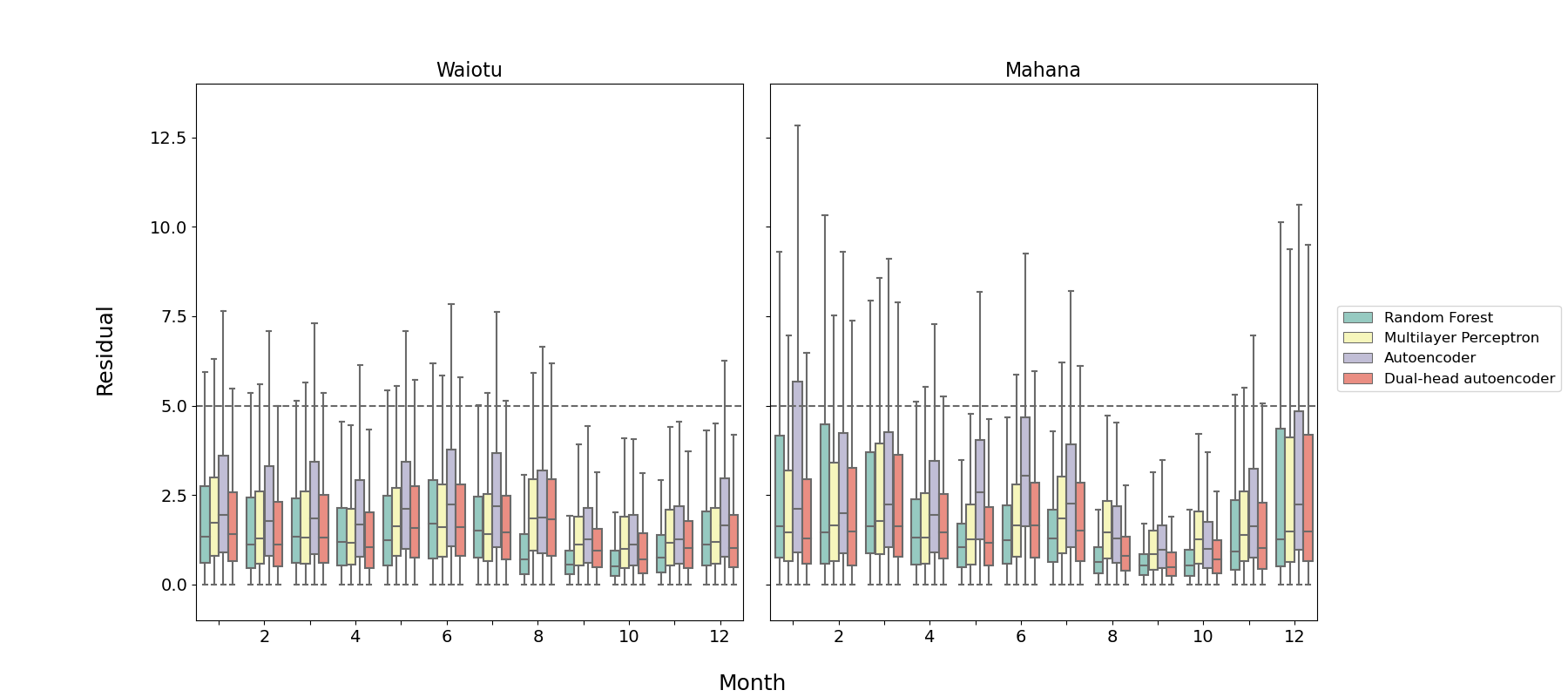}
\end{center}
\caption{Residuals for each architecture and RF aggregated over years and runs. The horizontal dashed line indicates the domain-derived threshold. The body of the candles represents $50\%$ of the values, and the bottom and top whiskers $25\%$ each. The horizontal lines inside the candles show the median.}
\label{fig:residuals}
\end{figure}

In Fig.~\ref{fig:residuals}, we can see how the residuals of the different architectures compared to RF across months. The residuals were aggregated over years and the five runs. For the first location, Waiotu, we observe that all candle bodies are below the domain-derived threshold that we set, with some upper whiskers overcoming the threshold. DAE seems to be the best performing architecture, since it has the shortest body of the three and also lower medians. Also, DAE appears to have slightly lower errors than RF in several cases. AE appears to have the largest errors, with large candles and extended upper whiskers. For Mahana, most candles are below our threshold but with larger bodies than Waiotu and taller upper whiskers. For January and December AE is above and close to the threshold respectively, generally having the highest errors. MLP and DAE seem to outperform RF for January, February, and December. Again, DAE has the best performance of the three architectures with candles being lower than the rest and lower medians.

\section{Discussion}

From a performance-oriented perspective, we could deduce that RF is the best model by looking at the error metrics. However, we cannot judge how much better it is from MLP and DAE or how well the different architectures learned because their errors and standard deviations were similar. A more clear case is that of AE, which underperforms the rest of the models considerably. The standard deviation of its predictions might be the smallest but this may be due to learning a small part of the lower dimensional manifold, created on the output of the encoder, and thus not being able to offer varied predictions. 

Examining the residuals of the architectures, we observe that they provide predictions mostly within our domain-derived threshold. The multiple runs and yearly aggregation demonstrate the stability\footnote{Variation between the months exists due to seasonality. December to February is summer in New Zealand with conditions that increase uncertainty for pasture growth and thus NRR errors.} of the models, showcasing their robustness. This conveys that the models were able to extract latent representations which allow them to be potentially used in an operational setting. AE appeared to be the weakest model since its candles were generally larger, exceeding our threshold in Mahana. The two stage training (first autoencoder, then replacing then decoder with an MLP) may have caused it to weigh more on learning how to reconstruct its inputs rather than how NRR is connected with them. On the other hand, the MLP was able to extract more meaningful representations for NRR predictions something evident from the fact that in many months it was on par and sometimes better than RF. Similarly, DAE performed equal or better than RF in most months for both locations. This may imply that the latent space that MLP and DAE learned covered aspects which were not represented in the manually derived expert features of RF. Also, the performance gap between AE and DAE showed that optimizing simultaneously for two tasks when one task depends on the other can make the network learn better representations in the context of this study.

An aspect potentially affecting the results of the architectures is how well input features can be represented in the latent space learned by the models. APSIM has a binary input variable to control the existence of irrigation which materially changes NRR. This variable is the only signal outside of APSIM that indicates this type of change. In our architectures there are several layers and this signal may be difficult to be preserved and projected in the latent space. On the contrary, for algorithms like RF this signal is not lost and can easily change how predictions are made. This may be a reason for not having higher performance with the different tested architectures and something to be accounted for when learning latent representations from environmental data.

\section{Conclusion and Future work}

In this study, we assessed the ability of three neural network architectures to learn the latent space of process-based model output for operational decision support. We compared the results with those of RF which was already proved operational in another study. The results were promising since all architectures were able to learn representations that captured enough variation to be considered operational. The MLP and DAE outperformed RF in certain cases, showing that they can uncover latent factors from the input space which accounted for more variability than manually selected features based on domain knowledge. This is an important step towards providing operational decision support in modern systems like digital twins, avoiding feature engineering in certain cases and automating prediction pipelines.

In the future, we would like to experiment with more synthetic datasets to examine if we can generalize our findings to other case studies. Also, we would like to validate the models using observation data to further verify how operational the created models are. Another important aspect would be to experiment with architectures that provide explicit interpretability of the latent space and examine how this space compares with expert-derived features. 

\subsubsection*{Acknowledgments}
This work has been supported by the European Union Horizon 2020 Research and Innovation program (Grant \#810775, ``Dragon'') and the Wageningen University and Research Investment Program ``Digital Twins''.

\bibliography{references}
\bibliographystyle{iclr2022_conference}

\pagebreak
\section*{Appendix}
\appendix

\section{Case study sites}

\begin{figure}[htp]
\begin{center}
\includegraphics[width=0.45\linewidth]{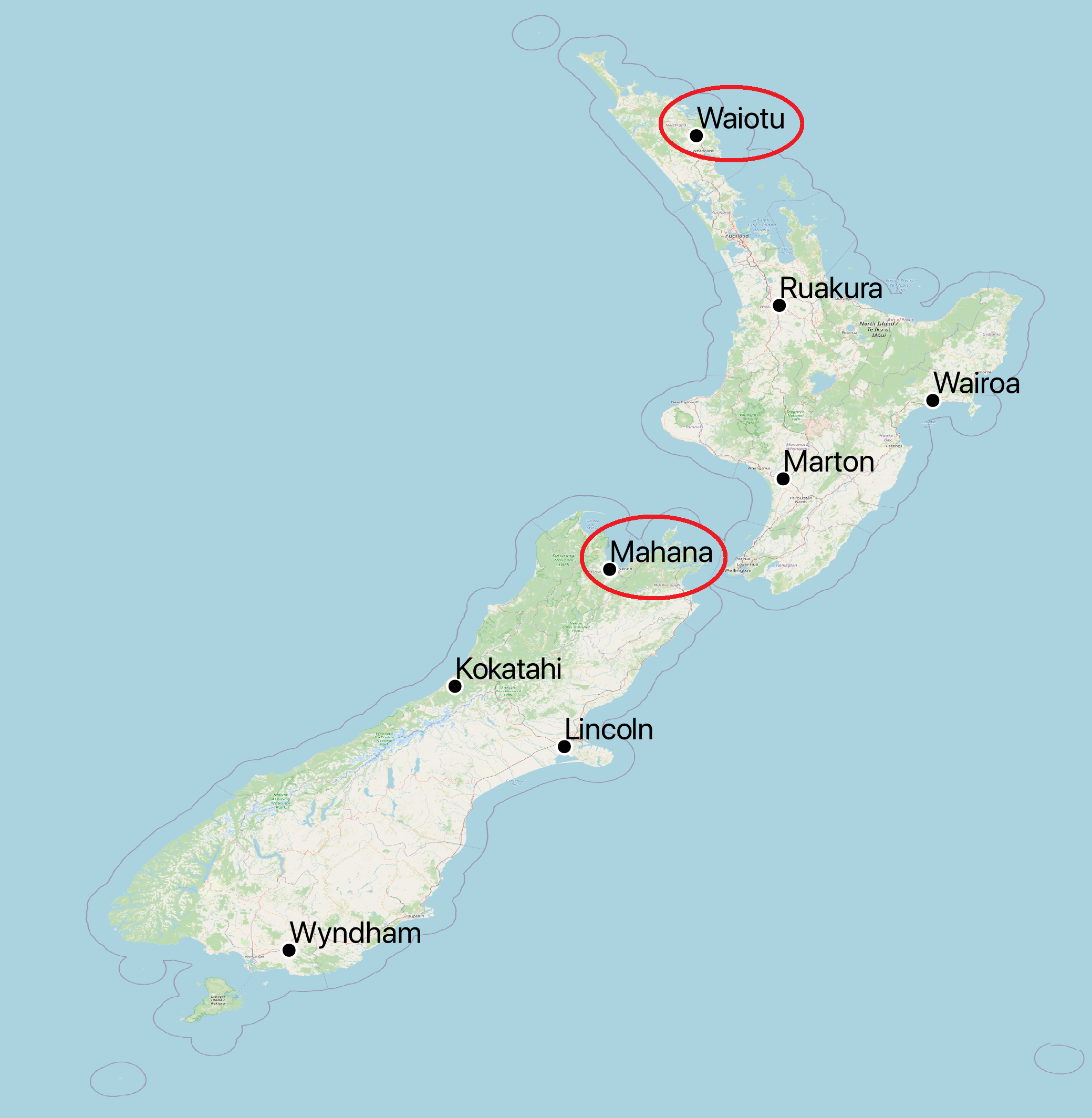}
\end{center}
\caption{New Zealand sites. Sites on the red circles are the ones included in this work.}
\label{fig:NZ_sites}
\end{figure}

\section{APSIM simulation parameters}

\begin{table}[htp]
\caption{APSIM simulation parameters and their ranges. The full factorial of those parameters comprised the input to APSIM.}
\label{tab:apsim_inputs}
\begin{center}
\begin{tabular}{ll}
\multicolumn{1}{l}{\bf Parameter}  &\multicolumn{1}{l}{\bf Range}
\\ \hline \\
    Weather & daily weather from 3 sites  \\
    Soil water & 42, 67, 110 and 177 mm of plant-available water\\
    Soil fertility & 2, 4, and 6\% of carbon concentration\\
    Irrigation & irrigated, non-irrigated \\
    Fertilizer year &  1979-2018 \\
    Fertilizer month & January-December \\
    Fertilizer day & $5^{th}$, $15^{th}$ and $25^{th}$ of the month \\
    Fertilizer amount & 0, 20, 40, 60, 80 and 100 kg N / ha \\
\end{tabular}
\end{center}
\end{table}

\section{Tuning and training}\label{tuning_training}

The training data were standardized for each location independently. The test and validation data were standardized with the corresponding training scaler, to have the same mean and standard deviation. 

The number of layers, nodes in each layer, optimizer parameters, and dropout rate for each architecture were based on the results of a preliminary study. The MLP had two hidden layers with 480 nodes each, optimization with Adam (lr=0.001, weight\_decay=0.0001), dropout rate 20\%, batch size 64 and 100 epochs. The AE had five hidden layers (300, 200, 120, 200, 300 nodes), optimization with AdamW (lr=0.0003, weight\_decay=0.01), dropout rate 10\%, batch size 64 and 60 epochs. After training, the decoder was replaced with an MLP with two hidden layers of 180 nodes each and training for 60 epochs. The DAE had the same autoencoder and optimizer as AE, with an addition of an MLP connected to the output of the encoder. The MLP had two hidden layers (80, 40 nodes). The whole network was trained with batch size 64, for 100 epochs.

RF took as input weekly aggregated features which were only a few and were considered explanatory so no feature selection took place. Hyperparameter tuning was performed using Bayesian optimization with 25 iterations and the 5-fold cross-validation score as a metric for each iteration. The tuned parameters can be seen in Table~\ref{tab:rf_tuning_parameters}.

\begin{table}[htp]
\caption{The parameters tuned during Bayesian optimization for RF.}
\label{tab:rf_tuning_parameters}
\begin{center}
\begin{tabular}{lc}
\multicolumn{1}{c}{\bf Parameters}  &\multicolumn{1}{c}{\bf Range}
\\ \hline \\
    n\_estimators         &  {50-800}\\
    max\_depth            &  {3-12}\\
    min\_samples\_split   &  {30-500}\\
    min\_samples\_leaf    &  {30-500}\\
    max\_features         &  {0.33}\\
\end{tabular}
\end{center}
\end{table}

\section*{} 
\end{document}

%% file: math_commands.tex
\usepackage{amsmath,amsfonts,bm}

\def\eqref#1{equation~\ref{#1}}

\def\1{\bm{1}}

\DeclareMathAlphabet{\mathsfit}{\encodingdefault}{\sfdefault}{m}{sl}
\SetMathAlphabet{\mathsfit}{bold}{\encodingdefault}{\sfdefault}{bx}{n}













%% file: main.bbl
\begin{thebibliography}{20}
\providecommand{\natexlab}[1]{#1}
\providecommand{\url}[1]{\texttt{#1}}
\expandafter\ifx\csname urlstyle\endcsname\relax
  \providecommand{\doi}[1]{doi: #1}\else
  \providecommand{\doi}{doi: \begingroup \urlstyle{rm}\Url}\fi

\bibitem[Antony(2014)]{Antony20146Designs}
Jiju Antony.
\newblock {6 - Full Factorial Designs}.
\newblock In Jiju Antony (ed.), \emph{Design of Experiments for Engineers and
  Scientists (Second Edition)}, pp.\  63--85. Elsevier, Oxford, 2014.
\newblock ISBN 978-0-08-099417-8.
\newblock \doi{https://doi.org/10.1016/B978-0-08-099417-8.00006-7}.
\newblock URL
  \url{https://www.sciencedirect.com/science/article/pii/B9780080994178000067}.

\bibitem[Breiman(2001)]{Breiman2001RandomForests}
Leo Breiman.
\newblock {Random Forests}.
\newblock \emph{Machine Learning}, 45\penalty0 (1):\penalty0 5--32, 2001.
\newblock ISSN 1573-0565.
\newblock \doi{10.1023/A:1010933404324}.
\newblock URL \url{https://doi.org/10.1023/A:1010933404324}.

\bibitem[Ghosh et~al.(2021)Ghosh, Renganathan, Khandelwal, Jia, Li, Neiber,
  Duffy, and Kumar]{Ghosh2021Knowledge-guidedCharacteristics}
Rahul Ghosh, Arvind Renganathan, Ankush Khandelwal, Xiaowei Jia, Xiang Li, John
  Neiber, Chris Duffy, and Vipin Kumar.
\newblock {Knowledge-guided Self-supervised Learning for estimating River-Basin
  Characteristics}.
\newblock 9 2021.
\newblock URL \url{http://arxiv.org/abs/2109.06429}.

\bibitem[Han et~al.(2015)Han, Okamoto, Beatty, Rothstein, and
  Good]{Han2015ThePlants}
Mei Han, Mamoru Okamoto, Perrin~H Beatty, Steven~J Rothstein, and Allen~G Good.
\newblock {The Genetics of Nitrogen Use Efficiency in Crop Plants}.
\newblock \emph{Annual Review of Genetics}, 49\penalty0 (1):\penalty0 269--289,
  11 2015.
\newblock ISSN 0066-4197.
\newblock \doi{10.1146/annurev-genet-112414-055037}.
\newblock URL \url{https://doi.org/10.1146/annurev-genet-112414-055037}.

\bibitem[He et~al.(2016)He, Zhang, Ren, and Sun]{He2016DeepRecognition}
K~He, X~Zhang, S~Ren, and J~Sun.
\newblock {Deep Residual Learning for Image Recognition}.
\newblock In \emph{2016 IEEE Conference on Computer Vision and Pattern
  Recognition (CVPR)}, pp.\  770--778, 2016.
\newblock ISBN 1063-6919.
\newblock \doi{10.1109/CVPR.2016.90}.

\bibitem[Holzworth et~al.(2014)Holzworth, Huth, deVoil, Zurcher, Herrmann,
  McLean, Chenu, van Oosterom, Snow, Murphy, Moore, Brown, Whish, Verrall,
  Fainges, Bell, Peake, Poulton, Hochman, Thorburn, Gaydon, Dalgliesh,
  Rodriguez, Cox, Chapman, Doherty, Teixeira, Sharp, Cichota, Vogeler, Li,
  Wang, Hammer, Robertson, Dimes, Whitbread, Hunt, van Rees, McClelland,
  Carberry, Hargreaves, MacLeod, McDonald, Harsdorf, Wedgwood, and
  Keating]{Holzworth2014APSIMSimulation}
Dean~P. Holzworth, Neil~I. Huth, Peter~G. deVoil, Eric~J. Zurcher, Neville~I.
  Herrmann, Greg McLean, Karine Chenu, Erik~J. van Oosterom, Val Snow, Chris
  Murphy, Andrew~D. Moore, Hamish Brown, Jeremy~P.M. Whish, Shaun Verrall,
  Justin Fainges, Lindsay~W. Bell, Allan~S. Peake, Perry~L. Poulton, Zvi
  Hochman, Peter~J. Thorburn, Donald~S. Gaydon, Neal~P. Dalgliesh, Daniel
  Rodriguez, Howard Cox, Scott Chapman, Alastair Doherty, Edmar Teixeira,
  Joanna Sharp, Rogerio Cichota, Iris Vogeler, Frank~Y. Li, Enli Wang,
  Graeme~L. Hammer, Michael~J. Robertson, John~P. Dimes, Anthony~M. Whitbread,
  James Hunt, Harm van Rees, Tim McClelland, Peter~S. Carberry, John~N.G.
  Hargreaves, Neil MacLeod, Cam McDonald, Justin Harsdorf, Sara Wedgwood, and
  Brian~A. Keating.
\newblock {APSIM - Evolution towards a new generation of agricultural systems
  simulation}.
\newblock \emph{Environmental Modelling and Software}, 62:\penalty0 327--350,
  12 2014.
\newblock ISSN 13648152.
\newblock \doi{10.1016/j.envsoft.2014.07.009}.

\bibitem[Jean et~al.(2018)Jean, Wang, Samar, Azzari, Lobell, and
  Ermon]{Jean2018Tile2Vec:Data}
Neal Jean, Sherrie Wang, Anshul Samar, George Azzari, David Lobell, and Stefano
  Ermon.
\newblock {Tile2Vec: Unsupervised Representation Learning for Spatially
  Distributed Data}.
\newblock Technical report, 2018.
\newblock URL \url{www.aaai.org}.

\bibitem[Kopf \& Claassen(2021)Kopf and Claassen]{Kopf2021LatentMedicine}
Andreas Kopf and Manfred Claassen.
\newblock {Latent representation learning in biology and translational
  medicine}.
\newblock \emph{Patterns}, 2\penalty0 (3):\penalty0 100198, 2021.
\newblock ISSN 2666-3899.
\newblock \doi{https://doi.org/10.1016/j.patter.2021.100198}.
\newblock URL
  \url{https://www.sciencedirect.com/science/article/pii/S2666389921000015}.

\bibitem[Li et~al.(2018)Li, Fang, Wu, and Wang]{Li2018AutoencoderEstimation}
Lianfa Li, Ying Fang, Jun Wu, and Jinfeng Wang.
\newblock {Autoencoder Based Residual Deep Networks for Robust Regression
  Prediction and Spatiotemporal Estimation}, 2018.

\bibitem[Mai et~al.(2022)Mai, Xuan, Zuo, Janowicz, and
  Lao]{Mai2022Sphere2Vec:Predictions}
Gengchen Mai, Yao Xuan, Wenyun Zuo, Krzysztof Janowicz, and Ni~Lao.
\newblock {Sphere2Vec: Multi-Scale Representation Learning over a Spherical
  Surface for Geospatial Predictions}.
\newblock 1 2022.
\newblock URL \url{http://arxiv.org/abs/2201.10489}.

\bibitem[Nasirahmadi \& Hensel(2022)Nasirahmadi and
  Hensel]{Nasirahmadi2022TowardParadigm}
Abozar Nasirahmadi and Oliver Hensel.
\newblock {Toward the Next Generation of Digitalization in Agriculture Based on
  Digital Twin Paradigm}.
\newblock \emph{Sensors}, 22\penalty0 (2), 1 2022.
\newblock ISSN 14248220.
\newblock \doi{10.3390/s22020498}.

\bibitem[Neumann et~al.(2019)Neumann, Pinto, Zhai, and
  Houlsby]{Neumann2019In-domainSensing}
Maxim Neumann, Andre~Susano Pinto, Xiaohua Zhai, and Neil Houlsby.
\newblock {In-domain representation learning for remote sensing}.
\newblock 11 2019.
\newblock URL \url{http://arxiv.org/abs/1911.06721}.

\bibitem[Pylianidis et~al.(2022)Pylianidis, Snow, Overweg, Osinga, Kean, and
  Athanasiadis]{Pylianidis2022Simulation-assistedTwins}
Christos Pylianidis, Val Snow, Hiske Overweg, Sjoukje Osinga, John Kean, and
  Ioannis~N Athanasiadis.
\newblock {Simulation-assisted machine learning for operational digital twins}.
\newblock \emph{Environmental Modelling {\&} Software}, 148:\penalty0 105274,
  2022.
\newblock ISSN 1364-8152.
\newblock \doi{https://doi.org/10.1016/j.envsoft.2021.105274}.
\newblock URL
  \url{https://www.sciencedirect.com/science/article/pii/S1364815221003169}.

\bibitem[Rivai et~al.(2021)Rivai, Miyamoto, Awano, Takada, Tobimatsu, Umezawa,
  and Kobayashi]{Rivai2021NitrogenSeedlings}
Reza~Ramdan Rivai, Takuji Miyamoto, Tatsuya Awano, Rie Takada, Yuki Tobimatsu,
  Toshiaki Umezawa, and Masaru Kobayashi.
\newblock {Nitrogen deficiency results in changes to cell wall composition of
  sorghum seedlings}.
\newblock \emph{Scientific Reports}, 11\penalty0 (1):\penalty0 23309, 2021.
\newblock ISSN 2045-2322.
\newblock \doi{10.1038/s41598-021-02570-y}.
\newblock URL \url{https://doi.org/10.1038/s41598-021-02570-y}.

\bibitem[Rotz et~al.(2005)Rotz, Taube, Russelle, Oenema, Sanderson, and
  Wachendorf]{Rotz2005Whole-farmAgriculture}
C~A Rotz, F~Taube, M~P Russelle, J~Oenema, M~A Sanderson, and M~Wachendorf.
\newblock {Whole-farm perspectives of nutrient flows in grassland agriculture}.
\newblock \emph{Crop Science}, 45\penalty0 (6):\penalty0 2139--2159, 11 2005.
\newblock ISSN 0011-183X.
\newblock \doi{10.2135/cropsci2004.0523}.
\newblock URL
  \url{http://www.scopus.com/inward/record.url?scp=27644585386&partnerID=8YFLogxK
  http://www.scopus.com/inward/citedby.url?scp=27644585386&partnerID=8YFLogxK}.

\bibitem[Whitehead(1995)]{Whitehead1995GrasslandNitrogen}
David~Charles Whitehead.
\newblock \emph{{Grassland nitrogen}}.
\newblock CAB international, 1995.
\newblock ISBN 9780851989150.

\bibitem[Zhang et~al.(2015{\natexlab{a}})Zhang, Davidson, Mauzerall,
  Searchinger, Dumas, and Shen]{Zhang2015ManagingDevelopment}
Xin Zhang, Eric~A Davidson, Denise~L Mauzerall, Timothy~D Searchinger, Patrice
  Dumas, and Ye~Shen.
\newblock {Managing nitrogen for sustainable development}.
\newblock \emph{Nature}, 528\penalty0 (7580):\penalty0 51--59,
  2015{\natexlab{a}}.
\newblock ISSN 1476-4687.
\newblock \doi{10.1038/nature15743}.
\newblock URL \url{https://doi.org/10.1038/nature15743}.

\bibitem[Zhang et~al.(2015{\natexlab{b}})Zhang, Tan, Zhu, Yuan, Xie, and
  Sun]{Zhang2015TOND1Rice}
Yangjun Zhang, Lubin Tan, Zuofeng Zhu, Lixing Yuan, Daoxin Xie, and Chuanqing
  Sun.
\newblock {TOND1 confers tolerance to nitrogen deficiency in rice}.
\newblock \emph{The Plant Journal}, 81\penalty0 (3):\penalty0 367--376, 2
  2015{\natexlab{b}}.
\newblock ISSN 0960-7412.
\newblock \doi{https://doi.org/10.1111/tpj.12736}.
\newblock URL \url{https://doi.org/10.1111/tpj.12736}.

\bibitem[Zhou et~al.(2019{\natexlab{a}})Zhou, Le, Hua, He, and
  Mao]{Zhou2019ImagingTreatments}
Chunyan Zhou, Jing Le, Dengxin Hua, Tingyao He, and Jiandong Mao.
\newblock {Imaging analysis of chlorophyll fluorescence induction for
  monitoring plant water and nitrogen treatments}.
\newblock \emph{Measurement}, 136:\penalty0 478--486, 2019{\natexlab{a}}.
\newblock ISSN 0263-2241.
\newblock \doi{https://doi.org/10.1016/j.measurement.2018.12.088}.
\newblock URL
  \url{https://www.sciencedirect.com/science/article/pii/S026322411831234X}.

\bibitem[Zhou et~al.(2019{\natexlab{b}})Zhou, Liu, Thung, and
  Shen]{Zhou2019LatentData}
T~Zhou, M~Liu, K~H. Thung, and D~Shen.
\newblock {Latent Representation Learning for Alzheimer’s Disease Diagnosis
  With Incomplete Multi-Modality Neuroimaging and Genetic Data}.
\newblock \emph{IEEE Transactions on Medical Imaging}, 38\penalty0
  (10):\penalty0 2411--2422, 2019{\natexlab{b}}.
\newblock ISSN 1558-254X.
\newblock \doi{10.1109/TMI.2019.2913158}.

\end{thebibliography}
